# Deep learning for predicting refractive error from retinal fundus images

Avinash V. Varadarajan, MS[1*]
Ryan Poplin, MS[1*]
Katy Blumer, BS[1]
Christof Angermueller, PhD[1]
Joe Ledsam, MBChB[2]
Reena Chopra, BSc[3]
Pearse A. Keane, MD[2, 3]
Greg S. Corrado, PhD[1]
Lily Peng, MD, PhD[1**]
Dale R. Webster, PhD[1**]

*Equal contribution
**Equal contribution
[1]Google Research, Google Inc, Mountain View, CA, USA
[2]Google DeepMind, Google Inc, London, UK
[3]NIHR Biomedical Research Centre for Ophthalmology, Moorfields Eye Hospital NHS Foundation Trust and UCL Institute of Ophthalmology, United Kingdom

Corresponding Author:
Lily Peng, MD, PhD
Google Research
1600 Amphitheatre Way
Mountain View, CA 94043
lhpeng@google.com

**Abbreviations**

Age Related Eye Disease Study (AREDS)
Diopter (D)
Mean Absolute Error (MAE)
Refractive Error (RE)

**ABSTRACT**

Refractive error, one of the leading cause of visual impairment, can be corrected by simple interventions like prescribing eyeglasses. We trained a deep learning algorithm to predict refractive error from the fundus photographs from participants in the UK Biobank cohort, which were 45 degree field of view images and the AREDS clinical trial, which contained 30 degree field of view images. Our model used the "attention" method to identify features that are correlated with refractive error. Mean absolute error (MAE) of the algorithm's prediction compared to the refractive error obtained in the AREDS and UK Biobank. The resulting algorithm had a mean absolute error (MAE) of 0.56 diopters (95% CI: 0.55-0.56) for estimating spherical equivalent on the UK Biobank dataset and 0.91 diopters (95% CI: 0.89-0.92) for the AREDS dataset. The baseline expected MAE (obtained by simply predicting the mean of this population) was 1.81 diopters (95% CI: 1.79-1.84) for UK Biobank and 1.63 (95% CI: 1.60-1.67) for AREDS. Attention maps suggested that the foveal region was one of the most important areas used by the algorithm to make this prediction, though other regions also contribute to the prediction. The ability to estimate refractive error with high accuracy from retinal fundus photos has not been previously known and demonstrates that deep learning can be applied to make novel predictions from medical images. Given that several groups have recently shown that it is feasible to obtain retinal fundus photos using mobile phones and inexpensive attachments, this work may be particularly relevant in regions of the world where autorefractors may not be readily available.

## BACKGROUND

Uncorrected refractive error is one of the most common causes of visual impairment worldwide.[1] The prevalence of refractive error is increasing, particularly myopic errors in Western and Asian populations.[2] Although largely treatable with prescription spectacles or contact lenses, the vast majority of those affected by refractive error live in low-income countries with minimal access to eye care and therefore may not receive even this non-invasive treatment.[3]

Novel and portable instruments, such as smartphone attachments to image the fundus[4] or apps to measure visual acuity,[5] offer a low-cost method of screening and diagnosis eye disease in the developing world. They have shown promise in the assessment of diabetic retinopathy[6] and the optic nerve[7] but are limited by their requirement for expert graders to interpret the images.

Artificial intelligence (AI) has shown promising results in the diagnosis and interpretation of medical imaging. In particular a form of AI known as deep learning allows systems to learn predictive features directly from the images from a large dataset of labeled examples without specifying rules or features explicitly.[8] Recent applications of deep learning to medical imaging have produced systems with performance rivaling medical experts for detecting a variety of, including melanoma,[9] diabetic retinopathy,[10,11] and breast cancer lymph node metastases.[12,13] Deep learning can also characterize signals that medical experts can not typically extract from images alone, such as age, gender, blood pressure, and other cardiovascular health factors.[14]

In this study, we trained a deep learning model[15,16] to predict the refractive error from fundus images using two different datasets. We then used attention techniques to visualize and identify new image features associated with the ability to make predictions.

## METHODS

*Data sets*

We used two datasets in this study -- UK Biobank and AREDS. UK Biobank is an ongoing observational study that recruited 500,000 participants between 40-69 years old across the United Kingdom between 2006-2010. Each participant completed lifestyle questionnaires, underwent a series of health measurements, provided biological samples,[17] and were followed up for health outcomes. Approximately 70,000 participants underwent ophthalmological examination, which included an assessment of refractive error using autorefraction (RC-5000; Tomey Corporation, Nagoya, Japan) as well as paired non-mydriatric optical coherence tomography and 45-degree retinal fundus imaging using a Topcon 3D OCT 1000 Mark 2 device (Topcon Corporation, Tokyo, Japan). Participants who had undergone any eye surgery, including cataract surgery, were excluded from participating in the ophthalmological exams.

The Age-Related Eye Disease Study (AREDS) was a clinical trial in the United States that investigated the natural history and risk factors of age-related macular degeneration and cataracts. The trial enrolled participants between 1992-1998 and continued clinical follow-up until 2001 at 11 retinal specialty clinics. The study was approved by an independent data and safety monitoring committee and by the institutional review board for each clinical center. A total of 4,757 participants aged 55-80 at enrollment were followed for a median of 6.5 years.[18] As a part of an ophthalmological exam, the participants underwent subjective refraction as well as color fundus photography at baseline and at subsequent visits. Briefly, the protocol for refraction involved retinoscopy and then further refinement with subjective refraction.

Thirty-degree field color fundus photographs were acquired with a Zeiss FF-series camera (Carl Zeiss, Oberkochen, Germany) using a reading center-approved transparency film.[19] For each visit where refraction was performed, the corresponding macula-centered photos were used in this study.

A summary metric for refractive error, known as the ‘spherical equivalent’, can be calculated using the formula spherical power + 0.5*cylindrical power. Spherical equivalent was available for both the UK Biobank and AREDS dataset, but spherical power and cylindrical power were only available in the UK Biobank dataset.

Each dataset was split into a development set and a clinical validation set which was not accessed during model development (table 1).

*Development of the algorithm*

A deep neural network model is a sequence of mathematical operation often with millions of parameters (weights)[20] applied to input, such as pixel values in an image. Deep learning is the process of learning the right parameter values (“training”) such that this function performs a given task, such as generating a prediction from the pixels values in a retinal fundus photograph. TensorFlow[21], an open-source software library for deep learning, was used in the training and evaluation of the models.

The development dataset was divided into two parts: a “train” set and a “tune” set. The “tune” set is also commonly called the “validation” set, but to avoid confusion with a clinical validation set (which consists of data the the model did not train on), we are calling it the “tune” set. During the training process, the parameters of the neural network are initially set to random values. Then for each image, the prediction given by the model is compared to the known label from the training set and parameters of the model are then modified slightly to decrease the error on that image. This process, known as stochastic gradient descent, is repeated for every image in the training set until the model "learns" how to accurately compute the label from the pixel intensities of the image for all images in the training set. The tuning dataset was a random subset of the development dataset that was not used to train the model parameters, but was used as a small evaluation dataset for tuning the model. This tuning set comprised 10% of the UK Biobank dataset, and 11% of the AREDS dataset. With appropriate tuning and sufficient data, the resulting model was able to predict the labels (e.g., refractive error) on new images. In this study, we design a deep neural network that combines a ResNet[22] and a soft-attention[23] architecture (Figure 1). Briefly, the network consists of layers to reduce the size of the input image, three residual blocks[22] to learn predictive image features, a soft-attention layer[23] to select the most informative features, and two fully-connected layers to learn interactions between the selected features.

Prior to training, we applied an image quality filter to exclude images of poor quality, which excluded approximate 12% of the UK Biobank dataset. Because the vast majority of the AREDS images were of good quality, we did not exclude any of the AREDS images. We preprocessed the images for training and validation, and trained the neural network following the same procedure as in Gulshan et al.[10] We trained separate models to predict spherical power, cylindrical power, and spherical equivalent (Figure 1).

We used an early stopping criteria[24] to help avoid overfitting and to terminate training when the model performance (such as MAE) on a tuning dataset stopped improving. To further improve results, we averaged the results of 10 neural network models that were trained on the same data (ensembling[25]).

*Evaluating the algorithm*

We optimized for minimizing the mean absolute error (MAE) to evaluate model performance for predicting refractive error. We also calculated the R-squared value, but this was not used to select the operating points for model performance. In addition, to further characterize the performance of the algorithms, we examined how frequently the algorithms' predictions fell within a given error margin (see Statistical Analysis section).

*Statistical Analysis*

To assess the statistical significance of these results, we used the non-parametric bootstrap procedure: from the validation set of $N$ instances, sample $N$ instances with replacement and evaluate the model on this sample. By repeating this sampling and evaluation 2,000 times, we obtain a distribution of the performance metric (e.g. MAE), and report the 2.5 and 97.5 percentiles as 95% confidence intervals. We compared the algorithms' MAE to baseline accuracy, which was generated by calculating the MAE of the actual refractive error and the average refractive error.

To further assess statistical significance, we performed hypothesis testing using an one-tailed binomial test for the frequency of the model's prediction lying within several error margins for each prediction. The baseline accuracy (corresponding to the null hypothesis) was obtained by sliding a window of size equal to the error bounds (e.g. size 1 for ±0.5) across the population histogram, and taking the maximum of the summed histogram counts. This provides the maximum possible "random" accuracy (by guessing the center of the sliding window containing the maximum probability mass).

*Attention maps*

To visualize the most predictive eye features, we integrated a soft-attention layer into our network architecture. The layer takes as input image features learned by the preceding layers, predicts for each feature a weight that indicates its importance for making a prediction, and outputs the weighted average of image features. We generated individual attention maps of images by visualizing the predicted feature weights as a heatmap. We also generated aggregated attention maps by averaging predicted attention weights over multiple images.

## RESULTS

The baseline characteristics of the UK Biobank and AREDS cohorts are summarized in Table 1. Patients in the UK Biobank dataset were imaged once. Patients in the AREDS dataset was imaged multiple times during the course of the trial. The patients in the AREDS study were on average older than those in UK Biobank (mean age: 73.8 years in AREDS versus 56.8 years in UK Biobank). Hypermetropia was more common in the AREDS dataset. The distribution of sex and ethnicity were similar in the two groups.

Table 2 summarizes the performance of the model on the clinical validation sets from UK Biobank and AREDS. The model was trained jointly on both the UK Biobank and AREDS datasets to predict the spherical equivalent of the refractive error. Both UK Biobank and AREDS datasets reported spherical equivalent, but the individual spherical and cylindrical components were only available in the UK Biobank dataset. The MAE of the model on UK Biobank clinical validation data was 0.56D (95% CI: 0.55-0.56) and 0.91D (95% CI: 0.89-0.93) on the AREDS clinical validation dataset (See Table 2). The distribution of the predicted vs actual values for both datasets are visualized in Figure 2. The model's predicted values were within 1D of the actual values 86% of the time for the UK Biobank clinical validation set vs 50% for baseline accuracy. For AREDS, the model's prediction was within 1D 65% of the time vs 45% for baseline. The difference between the model and baseline were significant at all margins of error (Table 3).

We further trained separate models to predict the components of spherical equivalent -- spherical power and cylindrical power, using the UK Biobank dataset since these values were not available in the AREDS dataset. The model trained to predict the spherical component from retinal fundus images was quite accurate, with an MAE of 0.63D (95% CI: 0.63, 0.64), and $R^2$ of 0.88 (95% CI: 0.88, 0.89). In comparison, the model trained to predict cylindrical power was not very accurate, with an MAE of 0.43 (95% CI 0.42, 0.43), and $R^2$ of 0.05 (95% CI 0.04, 0.05) (See Table 4).

Attention maps were generated to visualize the regions on the fundus that were most important for the refractive error prediction. Representative examples of attention maps at different categories of severities of refractive error (myopia, hyperopia) are shown in Figure 3. For every image, the macula was a prominent feature that was highlighted. In addition, diffuse signals such as retinal vessels and cracks in retinal pigment were also highlighted. There was not an obvious difference in the heatmaps for different severities of refractive error. We averaged and merged the attention maps for 1000 images at different severities of refractive error and found that these observations also generalized across many images (Figures 4 and 5). Given the importance of the fovea in the model predictions, we also investigated the effect that eye disease may have on the accuracy of predictions. The UK Biobank dataset contained mostly healthy eyes and so could not be used for this analysis. Using the AREDS dataset, we subdivided the patient population based upon whether or not that the patient had cataracts and/or AMD. We found a small but significant improvement in the accuracy of the model when we excluded patients with cataracts and/or AMD from the analysis (Table 5).

## DISCUSSION

In this study, we have shown that deep learning models can be trained to predict refractive error from retinal fundus images with high accuracy, a surprising result given that this was not a prediction task thought to be possible from retinal fundus images. Attention maps, which highlight features predictive for refractive error, show a clear focus of attention to the fovea for all refractive errors. To the best of our knowledge, there is no prior literature exploring the relationship between the fovea imaged using a fundus camera and refractive error. Previous work with higher resolution using optical coherence tomography has shown some evidence for anatomical difference in the retinal thickness or contour at the fovea with varying refractive error.[26] Although there is some evidence for greater spacing of foveal cone photoreceptors in myopic eyes,[27] this is unlikely to be resolved in retinal fundus images. However we can only speculate on the reasons why the fovea is of importance to the prediction. Refractive error correction is based upon focusing light sharply onto the fovea. Is there a variation in the foveal light reflex captured by fundus photography with refractive error? Newcomb and Potter found no association between refractive error and the presence foveal light reflex when visualized using an ophthalmoscope - a feature that becomes dimmer and less conspicuous with increasing age[28] or presence of macular disease. The ‘brightness’ of this reflex and its relationship with refractive error has not been studied. Could there be a relationship between colour or macular pigment at the fovea and refractive error? Czepita et al. found no association between macular pigment density and refractive error.[29] The density of pigment is usually derived using psychophysical techniques but fundus photographs captured using blue and green illumination have shown promise in evaluating density.[30]

The attention maps also suggest that features outside the foveal region contribute to the prediction to a lesser extent, including a diffuse signal from the optic disc and retinal temporal vessel arcades from their exit from the optic nerve as they traverse across the fundus. The extent of association between optic disc size and refractive error is unresolved due to inconsistent findings among studies. Eyes with axial myopia may display tilted optic discs.[31] Some studies have shown a weakly significant increase in optic disc size with increasing refractive error towards myopia,[32,33] whereas a Chinese population-based study found that the optic disc size is independent of refractive error within the range of -8 to +4 diopters.[34] Varma et al. found no association between refractive error and optic disc size.[35] Myopic refractive errors have also been associated with narrower retinal arterioles and venules and increased branching,[36] and reduction in retinal vascular fractal dimensions.[37] In addition, the maps also looked very similar for images with hypermetria and myopia, suggesting that the neural network is leveraging the same regions for predictions over a spectrum of refractive errors.

While attention maps show anatomic correlates for the prediction of interest, they do not establish causation. This is a general limitation of existing attention techniques. However, these maps may be a way to generate hypotheses in a non-biased manner to further research into the pathophysiology of refractive error.

We found that the MAE of our joint model on the Biobank dataset was lower than on the AREDS dataset, which may be due to a variety of factors. Firstly, the camera used to image the fundus in the UK Biobank study was a wider 45-degree field camera which captured more peripheral information than the 30-degree field of the Zeiss camera used in the AREDS dataset. This results in the optic disc or retinal vessels (shown to be important in the UK Biobank model) not always being visible in the acquired image. Secondly, the AREDS dataset has far fewer images, and small training sets generally result in a decrease in generalizability and performance in the clinical validation set. In addition, many images in the AREDS dataset exhibited macular pathology of some form. Given the importance of the foveal region in prediction refractive error, this might have decreased the model's performance on the AREDS dataset. Thirdly, the refractive error was determined by two different methods in each dataset: autorefraction in the Biobank dataset versus subjective refraction in AREDS. We believe that the smaller capture field, preexisting eye pathologies, and smaller dataset combined lower the predictive power in the AREDS dataset relative to the UK Biobank.

The model has high accuracy when predicting spherical power, which is generally related to the axial length of the eye, than when predicting cylindrical power. This is expected as astigmatism is the result of toricity of the cornea and/or the crystalline lens, information that is unlikely to be held in retinal fundus images. Unfortunately, axial length data was unavailable for either dataset to confirm this hypothesis.

Fundus camera optics combined with the subject's refractive error are known to influence the optical magnification of retinal photographs.[38] In this study, we did not correct the images for ocular magnification as this requires application of a formula with known spherical equivalent refraction. Although there is a relationship between magnification and ametropia, it is improbable that the model is identifying differences in magnification over the range of refractive errors. Magnification correction is of greatest importance for analysis of features such as blood vessel parameters at patient-level, but its influence is likely to be minimal in large datasets as used in this study.

There are a few additional limitations for this study. Our algorithm can predict the spherical component of refractive error fairly well, but it does not do as well for the cylindrical component. The model was trained and validated on a combination of two datasets. It would be more desirable to have a third dataset that was taken in a completely different setting for additional validation. Future studies should include datasets from even more diverse populations, such as different ethnicities, ages, and comorbidities. There may also be additional application of this work in epidemiological (e.g. predict refractive error from unlabelled fundus image datasets) or pathophysiological studies of refractive error (e.g. using attention maps to for the causes and pathophysiology of myopia progression).

Currently autorefraction is no more difficult to perform than fundus photography, so the findings of this study are unlikely to change the role of autorefraction in most clinical settings. However, portable fundus cameras such as PEEK[39] are becoming less expensive and more common for screening and diagnosis of eye disease, particularly in the developing world. With further validation, it may be possible to use these increasingly abundant fundus images to efficiently screen for individuals with uncorrected refractive error who would benefit from a formal refraction assessment.

**ACKNOWLEDGEMENTS**

From Google Research: Mark DePristo, PhD, Arunachalam Narayanaswamy, PhD, Yun Liu, PhD

This research has been conducted using the UK Biobank Resource under Application Number 17643.

# FIGURES & TABLES

| | Development Set | | Clinical Validation Set | |
|---|---|---|---|---|
| **Characteristics** | **UK Biobank** | **AREDS** | **UK Biobank** | **AREDS** |
| Number of Patients | 48,101 | 4128 | 12,026 | 500 |
| Number of Images | 96,081 | 130,789 | 24,007 | 15,750 |
| Age at imaging visit(s): Mean, years (SD) | 56.8 (8.2) | 73.8 (4.92) | 56.9 (8.2) | 73.83 (5.22) |
| Sex (% male) | 44.9 | 44.3 | 44.9 | 42.8 |
| Ethnicity | 1.2% Black, 3.4% Asian/PI, 90.6% White, 4.1% Other 0.7%: Unknown | 3.7% Black, 0.2% Asian/PI, 95.7% White, 0.3% Hispanic, 0.2% Other | 1.3% Black, 3.6% Asian/PI, 90.1% White, 4.2% Other 0.8%: Unknown | 4.0% Black, 0.2% Asian/PI, 95.2% White, 0.4% Hispanic, 0.2% Other |
| Spherical Equivalent (SE): Mean, diopters (SD) | -0.38 (2.63) | 0.67 (2.00) | -0.34 (2.57) | 0.60 (2.08) |
| Severe Myopia; (SE worse than -6.00D): (%) | 3.9% | 0.7% | 3.8% | 0.6% |
| Moderate Myopia; (SE -3.00D to -6.00D): (%) | 9.6% | 4.0% | 9.2% | 5.4% |
| Mild Myopia; (SE up to -3.00D): (%) | 33.7% | 24.1% | 33.5% | 23.8% |
| Mild Hypermetropia; (SE up to +2.00D): (%) | 41.1% | 50.1% | 41.7% | 47.0% |
| Moderate Hypermetropia; (SE +2.00 to +5.00): (%) | 9.6% | 19.9% | 9.8% | 21.8% |
| Severe Hypermetropia; (SE worse than +5.00): (%) | 1.3% | 1.1% | 1.2% | 1.4% |
| Unknown SE: (%) | 0.8% | 0.0% | 0.8% | 0.0% |

**Table 1:** Population characteristics of patients in the UK Biobank and AREDS datasets. The spherical equivalent (SE) values shown are the averaged value over boths eyes in the case of the UK Biobank dataset, and the averaged value over both eyes across all the visits in the AREDS data set.

| **Dataset** | **MAE** | | **R2** | |
|---|---|---|---|---|
| | Model | Baseline | Model | Baseline |
| UK Biobank (n=23520) | 0.56 [0.55, 0.56] | 1.81 [1.79-1.84] | 0.90 [0.90, 0.91] | 0.0 [0.0, 0.0] |
| AREDS (n=7635) | 0.91 [0.89, 0.93] | 1.63 [1.60-1.67] | 0.69 [0.66, 0.71] | 0.0 [0.0, 0.0] |

**Table 2:** Mean absolute error (MAE) and coefficient of determination (R2) of algorithm vs baseline for predicting the spherical equivalent. Baseline metrics are calculated by predicting mean values of the validations set. The 95% confidence intervals are shown in square brackets; all the values are in units of diopters.

| | **UK Biobank Validation Set** (n=23,520) | | | **AREDS Validation Set** (n=7,635) | | |
|---|---|---|---|---|---|---|
| Error Margin (Diopters) | Model Accuracy | Baseline Accuracy* | p-value | Model Accuracy | Baseline Accuracy* | p-value |
| ±0.5 | 59% | 30% | <0.0001 | 37% | 25% | <0.0001 |
| ±1 | 86% | 50% | <0.0001 | 65% | 45% | <0.0001 |
| ±2 | 97% | 71% | <0.0001 | 91% | 74% | <0.0001 |

**Table 3:** Model accuracy vs baseline for predicting spherical equivalent within a given margin.
*Baseline accuracy was generated by sliding a window of size equal to the error bounds (e.g. size 1 Diopter for ±0.5) across the population histogram, and taking the maximum of the summed histogram counts. This provides the maximum possible “random” accuracy (by guessing the center of the sliding window corresponding to the maximum)

| | **UK Biobank Validation Set** (n=23,520) | | | **AREDS Validation Set** (n=7,635) | | |
|---|---|---|---|---|---|---|
| Error Margin (Diopters) | Model Accuracy | Baseline Accuracy* | p-value | Model Accuracy | Baseline Accuracy* | p-value |
| ±0.5 | 59% | 30% | <0.0001 | 37% | 25% | <0.0001 |
| ±1 | 86% | 50% | <0.0001 | 65% | 45% | <0.0001 |
| ±2 | 97% | 71% | <0.0001 | 91% | 74% | <0.0001 |

**Table 3:** Model accuracy vs baseline for predicting spherical equivalent within a given margin. *Baseline accuracy was generated by sliding a window of size equal to the error bounds (e.g. size 1 Diopter for ±0.5) across the population histogram, and taking the maximum of the summed histogram counts. This provides the maximum possible “random” accuracy (by guessing the center of the sliding window corresponding to the maximum)

| | **MAE** | | **R2** | |
|---|---|---|---|---|
| | Model | Baseline | Model | Baseline |
| **Cylindrical Component** UK Biobank | 0.43 [0.42, 0.43] | 0.48 [0.47-0.49] | 0.05 [0.04, 0.06] | 0.0 [0.0, 0.0] |
| **Spherical Component** UK Biobank | 0.63 [0.63, 0.64] | 1.89 [1.87-1.92] | 0.88 [0.88, 0.89] | 0.0 [0.0, 0.0] |

**Table 4:** Mean absolute error (MAE) and coefficient of determination (R2) of algorithm vs baseline for predicting the Cylindrical and Spherical components of the refractive error in the UK Biobank dataset . Baseline metrics are calculated by predicting mean values of the validations set. The 95% confidence intervals are shown in square brackets; all the values are in units of diopters.

| | **MAE** | **R-squared** |
|---|---|---|
| **All Patients** (n=7,635) | 0.91 [0.89, 0.93] | 0.69 [0.66, 0.71] |
| **No Cataract Surgery** (n=5,071) | 0.89 [0.87, 0.91] | 0.73 [0.71, 0.74] |
| **No AMD** (n=2,410) | 0.84 [0.81, 0.87] | 0.79 [0.77, 0.80] |
| **No Cataract Surgery or AMD** | 0.82 | 0.80 |
| (n=1,855) | [0.79, 0.85] | [0.78, 0.81] |

**Table 5:** Performance of the Algorithm on the AREDS validation set sliced by AMD diagnosis and history of Cataract Surgery

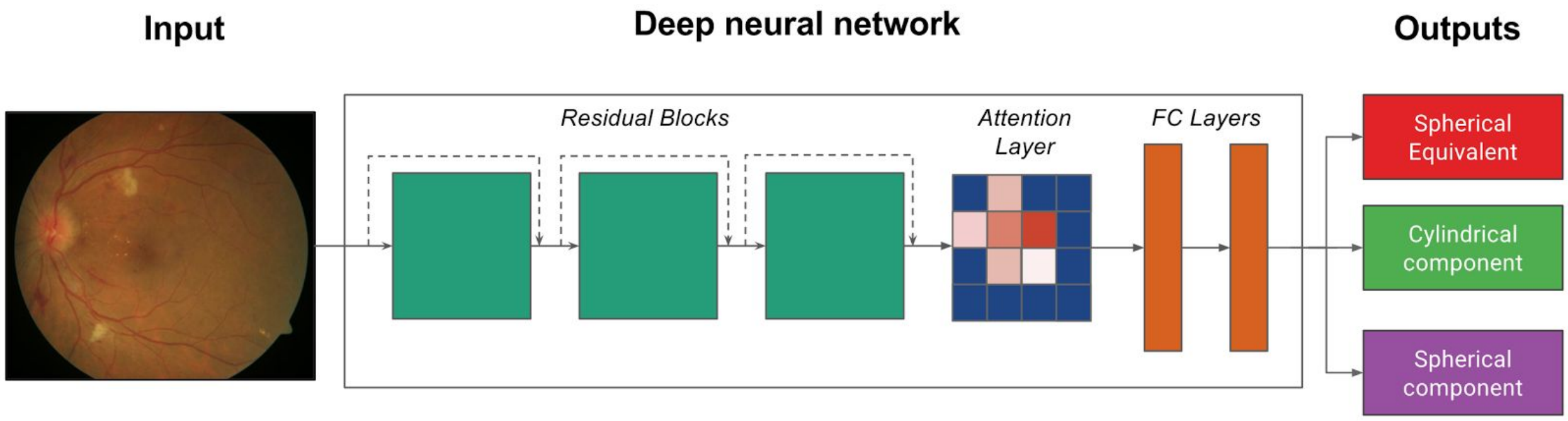

**FIgure 1.** Overview diagram. Fundus images form the input of a deep neural network consisting of three residual blocks, an attention layer to learn the most predictive eye features, and two fully connected layers. Model outputs are spherical equivalent, cylindrical component, and spherical component. Model parameters are learnt in a data-driven manner by showing input-output examples.

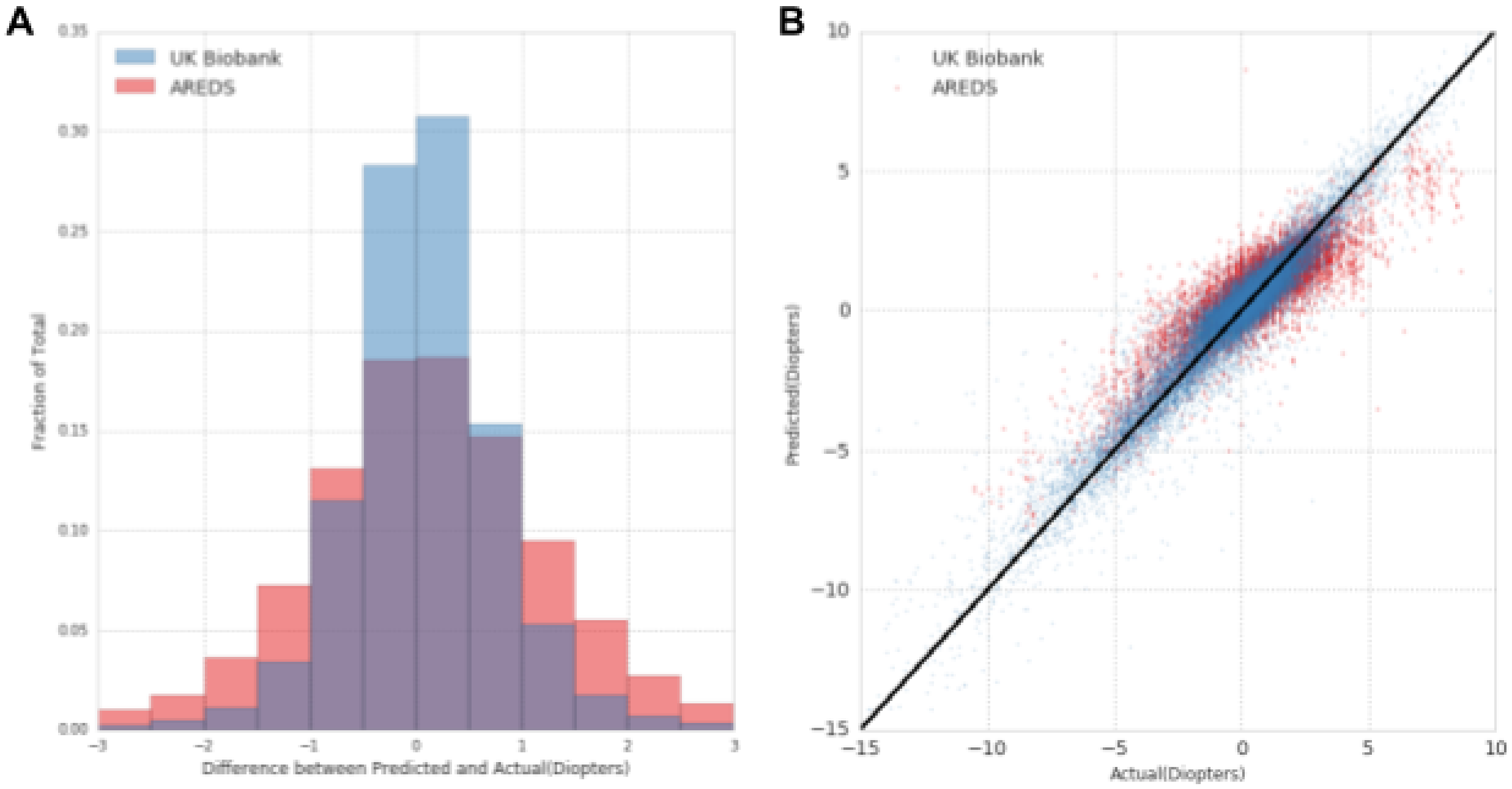

**Figure 2**. Model performance in predicting spherical equivalent on the two clinical validation sets. (A) Histogram of prediction error (Predicted - Actual) UK Biobank (blue) and AREDS dataset (red). (B)

Scatter plot of predicted and actual values for each instance in the validation sets. Black diagonal indicates perfect prediction, where y=x.

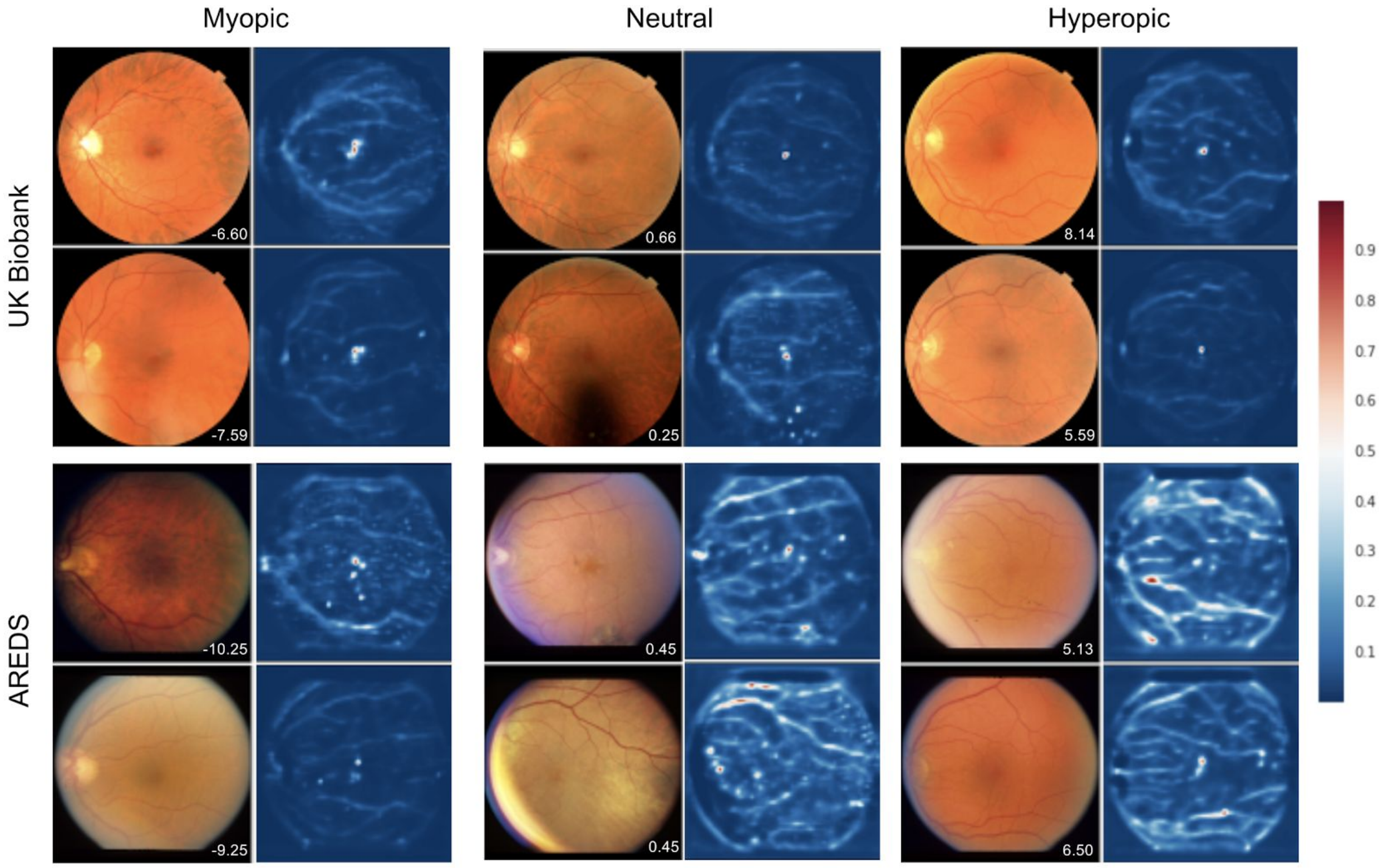

**Figure 3.** Example attention maps for three left myopic (SE worse than -6.0), neutral (SE between -1.0 and 1.0), and hyperopic (SE worse than 5.0) fundus images from UK Biobank (two top rows) and AREDS (two bottom rows). Diagnosed spherical equivalent is printed in the bottom right corner of fundus images.

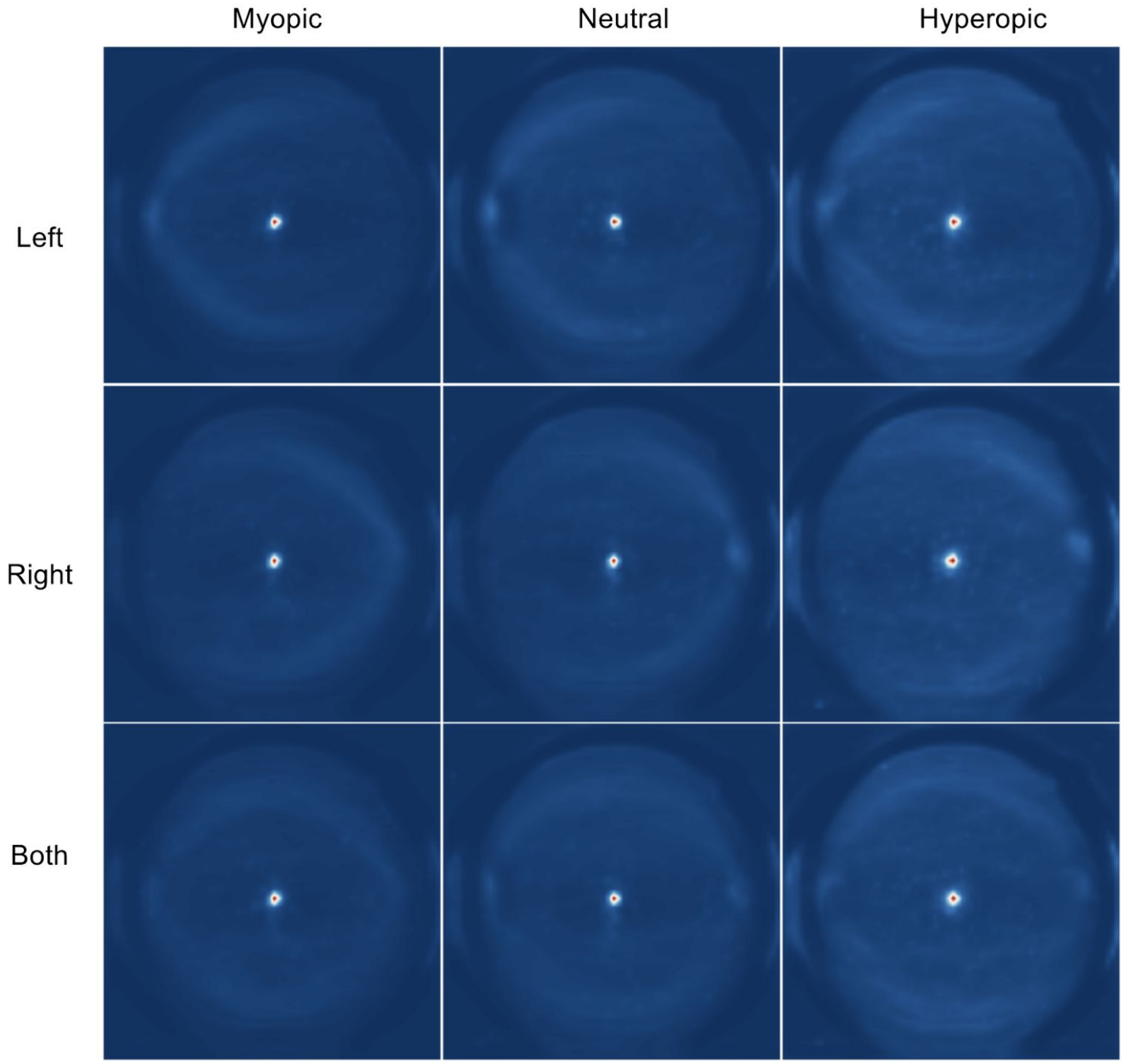

**Figure 4**. Mean attention map over 1000 images from UK Biobank for severely myopic (SE worse than -6.0), neutral (SE between -0.5 and 0.5), and severely hyperopic (SE worse than 5.0) eyes conditioned on eye position.

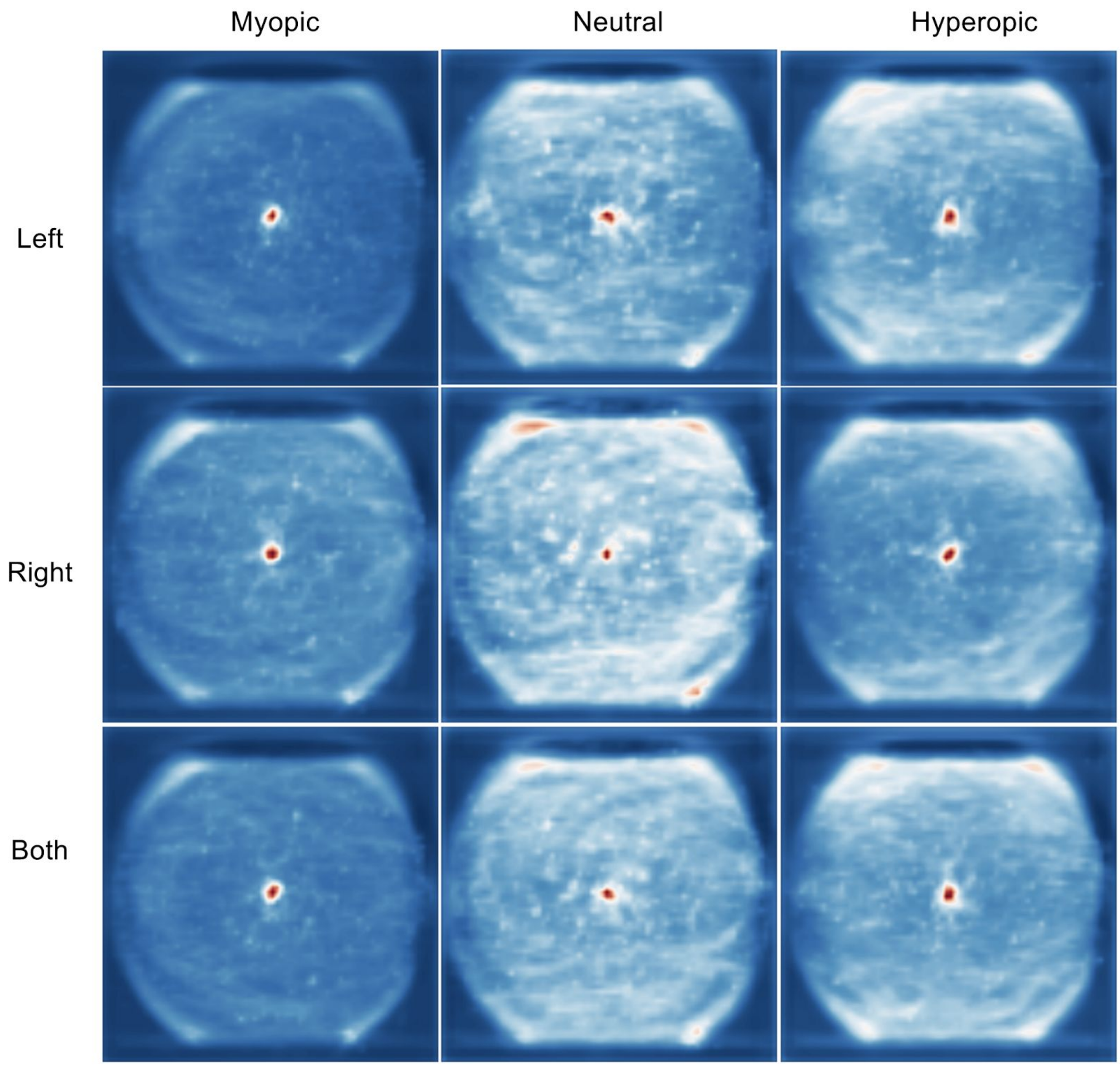

**Figure 5**. Mean attention map over 1000 images from AREDS for severely myopic (SE worse than -6.0), neutral (SE between -0.5 and 0.5), and severely hyperopic (SE worse than 5.0) eyes conditioned on eye position.

(Accessed: 4th December 2017)